\documentclass{article} 
\usepackage{iclr2016_workshop,times}
\usepackage{hyperref}
\usepackage{url}
\usepackage{bbm}
\usepackage{graphicx}
\usepackage{enumerate}
\usepackage{subcaption}
\usepackage{titlesec}

\usepackage{amsmath,amsfonts,amssymb}
\newcommand\transp{^\intercal\kern-\scriptspace}

\renewcommand{\vec}[1]{\mathbf{#1}}  

\DeclareMathOperator*{\argmax}{arg\,max}

\newenvironment{tightitemize} 
{\vspace{-\topsep}\begin{itemize}\itemsep1pt \parskip0pt \parsep0pt}
{\end{itemize}\vspace{-\topsep}}

\title{Stacked What-Where Auto-encoders}

\author{
Junbo Zhao, Michael Mathieu, Ross Goroshin, Yann LeCun \\
Courant Institute of Mathematical Sciences, New York University \\
719 Broadway, 12th Floor, New York, NY 10003 \\
\texttt{\{junbo.zhao, mathieu, goroshin, yann\}@cs.nyu.edu}
}

\begin{document}

\maketitle

\begin{abstract}
We present a novel architecture, the ``stacked what-where auto-encoders'' (SWWAE), which integrates discriminative and generative pathways and provides a unified approach to supervised, semi-supervised and unsupervised learning without relying on sampling during training. An instantiation of SWWAE uses a convolutional net (Convnet) (\cite{lecun1998gradient}) to encode the input, and employs a deconvolutional net (Deconvnet) (\cite{zeiler2010deconvolutional}) to produce the reconstruction. The objective function includes reconstruction terms that induce the hidden states in the Deconvnet to be similar to those of the Convnet. Each pooling layer produces two sets of variables: the ``what'' which are fed to the next layer, and its complementary variable ``where'' that are fed to the corresponding layer in the generative decoder.
\end{abstract}

\section{Introduction}
\label{sec:intro}
A desirable property of learning models is the ability to be trained in supervised, unsupervised, or semi-supervised mode with a single architecture and a single learning procedure. Another desirable property is the ability to exploit the advantageous discriminative and generative models.
A popular approach is to pre-train auto-encoders in a layer-wise fashion, and subsequently fine-tune the entire stack of encoders (the feed-forward pathway) in a supervised discriminative manner (\cite{erhan2010does,gregor-icml-10, henaff-ismir-11, koray-cvpr-09, koray-psd-08, koray-nips-10, ranzato2007unsupervised, ranzato2007sparse}). This approach fails to provide a unified mechanism to unsupervised and supervised learning. Another approach, that provides a unified framework for all three training modalities, is the deep boltzmann machine (DBM) model~(\cite{hinton2006fast, larochelle2008classification}). Each layer in a DBM is an restricted boltzmann machine (RBM), which can be seen as a kind of auto-encoder. Deep RBMs have all the desirable properties,  however they exhibit poor convergence and mixing properties ultimately due to the reliance on sampling during training. The main issue with stacked auto-encoders is asymmetry. The mapping implemented by the feed-forward pathway is often many-to-one, for example mapping images to invariant features or to class labels. Conversely, the mapping implemented by the feed-back (generative) pathway is one-to-many, e.g. mapping class labels to image reconstructions. The common way to deal with this is to view the reconstruction mapping as probabilistic. This is the approach of RBMs and DBMs: the missing information that is required to generate an image from a category label is dreamed up by sampling. This sampling approach can lead to interesting visualizations, but is impractical for training large scale networks because it tends to produce highly noisy gradients.

If the mapping from input to output of the feed-forward pathway were one-to-one, the mappings in both directions would be well-defined functions and there would be no need for sampling while reconstructing. But if the internal representations are to possess good invariance properties, it is desirable that the mapping from one layer to the next be many-to-one. For example, in a Convnet, invariance is achieved through layers of max-pooling and subsampling.

Our model attempts to satisfy two objectives: (i)-to learn a factorized representation that encodes invariance and equivariance, (ii)-we want to leverage both labeled and unlabeled data to learn this representation in a unified framework.
The main idea of the approach we propose here is very simple: whenever a layer implements a many-to-one mapping, we compute a set of complementary variables that enable reconstruction. A schematic of our model is depicted in figure \ref{fig:model_pool} (b). In the max-pooling layers of Convnets, we view the position of the max-pooling ``switches'' as the complementary information necessary for reconstruction. The model we proposed consists of a feed-forward Convnet, coupled with a feed-back Deconvnet. Each stage in this architecture is what we call a ``what-where auto-encoder''. The encoder is a convolutional layer with ReLU followed by a max-pooling layer.
The output of the max-pooling is the ``what'' variable, which is fed to the next layer. The complementary variables are the max-pooling ``switch'' positions, which can be seen as the ``where'' variables. The ``what'' variables inform the next layer about the content with incomplete information about position, while the ``where'' variables inform the corresponding feed-back decoder about where interesting (dominant) features are located. The feed-back (generative) decoder reconstructs the input by ``unpooling'' the ``what'' using the ``where'', and running the result through a reconstructing convolutional layer. Such ``what-where'' convolutional auto-encoders can be stacked and trained jointly without requiring alternate optimization (\cite{zeiler2010deconvolutional}).
The reconstruction penalty at each layer constrains the hidden states of the feed-back pathway to be close to the hidden states of the feed-forward pathway. The system can be trained in purely supervised manner: the bottom input of the feed-forward pathway is given the input, the top layer of the feed-back pathway is given the desired output, and the weights of the decoders are updated to minimize the sum of the reconstruction costs. If only the top-level cost is used, the model reverts to purely supervised backprop. If the hidden layer reconstruction costs are used, the model can be seen as supervised with a reconstruction regularization. In unsupervised mode, the top-layer label output is left unconstrained, and simply copied from the output of the feed-forward pathway. The model becomes a stacked convolutional auto-encoder. As with boltzmann machines (BM), the underlying learning algorithm doesn't change between the supervised and unsupervised modes and we can switch between different learning modalities by clamping or unclamping certain variables. Our model is particularly suitable when one is faced with a large amount of unlabeled data and a relatively small amount of labeled data. The fact that no sampling (or contrastive divergence method) is required gives the model good scaling properties; it is essentially just backprop in a particular architecture.

\section{Related work}
\label{sec:related}
The idea of ``what'' and ``where'' has been defined previously in different ways.
One related method was proposed known as ``transforming auto-encoders'' (\cite{hinton2011transforming}), in which ``capsule'' units were introduced. In that work, two sets of variables are trained to encapsulate ``invariance'' and ``equivariance'' respectively, by providing the parameters of particular transformation states to the network. Our work is carried out in a more unsupervised fashion in that it doesn't require the true latent state while still being able to encode similar representations within the ``what'' and ``where''.
Switches information is also made use of by some visualization work such as \cite{zeiler2010deconvolutional}, while such work only has a generative pass and merely uses a feed-forward pass as an initialization step.

Similar definitions have been applied to learn invariant features (\cite{gregor-icml-10, henaff-ismir-11, koray-cvpr-09, koray-psd-08, koray-nips-10, ranzato2007unsupervised, ranzato2007sparse, makhzani2014winner, masci2011stacked}). Among them, most works merely shed light
to unsupervised feature learning and therefore failed to unify different learning modalities. Another relevant hierarchical architecture is proposed in (\cite{ranzato2007unsupervised, ranzato2007sparse}), however, because this architecture is trained in a layer-wise greedy manner, its performance is not competitive with jointly trained models.

In terms of joint loss minimization and semi-supervised learning, our work can be linked to \cite{weston2012deep} and \cite{ranzato2008semi}, with the main advantage being the easiness to extend a Convnet with a Deconvnet and thereby enabling the utilization of unlabeled data. 
\cite{paine2014analysis} has analyzed the regularization effect with similar architectures in a layer-wise fashion.

One recent work (\cite{rasmus2015lateral}, \cite{rasmus2015semi}) has been proposed to adopt deep auto-encoders to support supervised learning in which completely different strategy is employed to harness the lateral connection between same stage encoder-decoder pairs, however. In that work, decoders receive the entire pre-pooled activation state from the encoder, whereas decoders from SWWAE only receive the ``where'' state from the corresponding encoder stages.
Further, due to a lack of unpooling mechanism incorporated in the Ladder networks, it is restricted to only reconstruct the top layer within generative pathway ($\Gamma$ model), which looses the "ladder" structure. By contrast, SWWAE doesn't suffer from such necessity.

\section{Model Architecture}
\label{sec:model}

\begin{figure}[h]
\centering
\includegraphics[width=0.7\columnwidth]{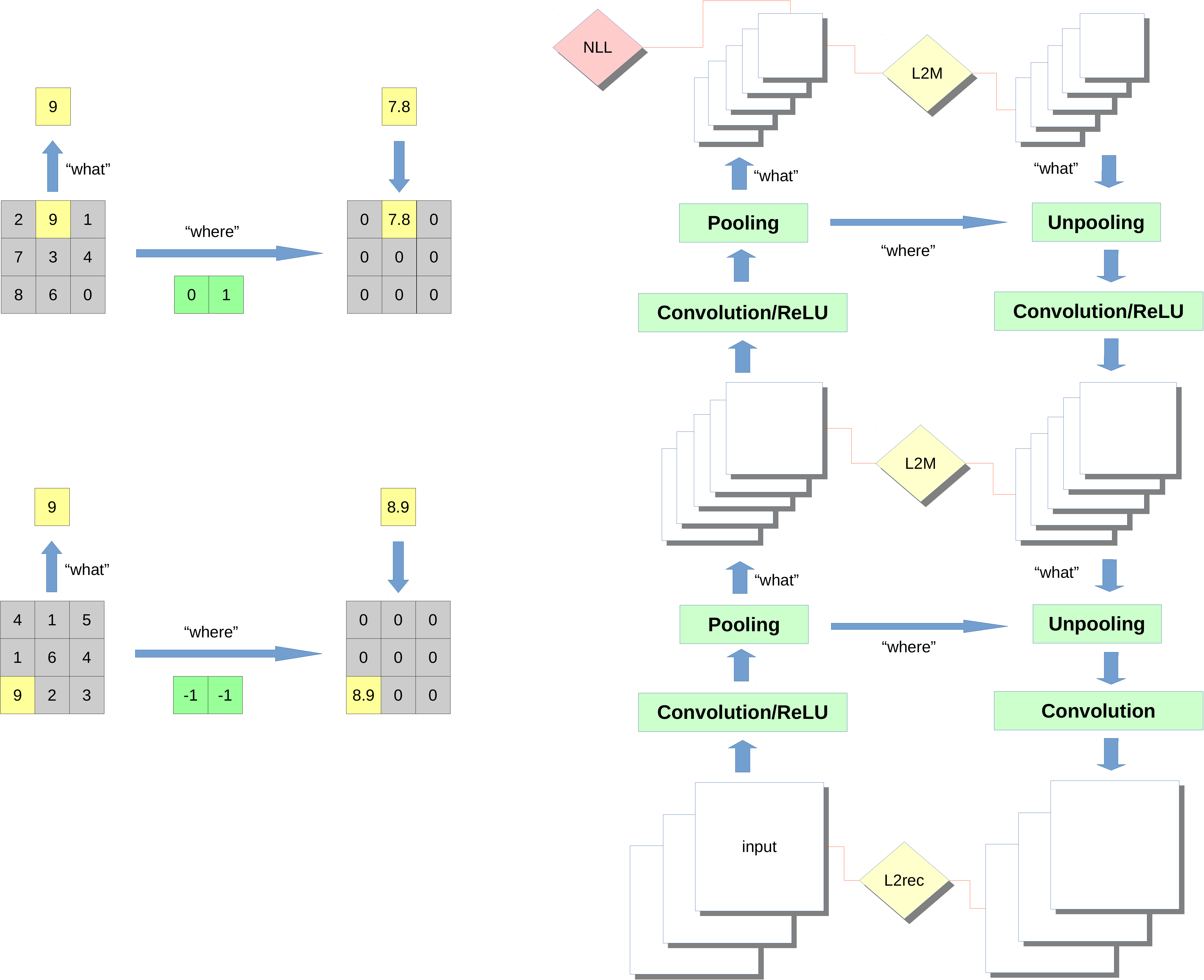}
\caption{Left (a): pooling-unpooling. Right (b): model architecture. For brevity, fully-connected layers are omitted in this figure.}
\label{fig:model_pool}
\end{figure}

We consider the loss function of SWWAE depicted in figure \ref{fig:model_pool}(b) composed of three parts:
\begin{equation}
L = L_{NLL} + \lambda_{L2rec}L_{L2rec} + \lambda_{L2M}L_{L2M},
\end{equation}
where $L_{NLL}$ is the discriminative loss, $L_{L2rec}$ is the reconstruction loss at the input level and $L_{L2M}$ charges intermediate reconstruction terms. $\lambda$'s weight the losses against each other.

Pooling layers in the encoder split information into ``what'' and ``where'' components, depicted in figure \ref{fig:model_pool}(a), that ``what'' is essentially \emph{max} and ``where'' carries \emph{argmax}, i.e., the switches of maximally activation defined under local coordinate frame over each pooling region. The ``what'' component is fed upward through the encoder, while the ``where'' is fed through lateral connections to the same stage in the feed-back decoding pathway.
The decoder uses convolution and ``unpooling'' operations to approximately invert the output of the encoder and reproduce the input, shown in figure \ref{fig:model_pool}. The unpooling layers use the ``where'' variables to unpool the feature maps by placing the ``what'' into the positions indicated the preserved switches. 
We use negative log-likelihood (NLL) loss for classification and L2 loss for reconstructions; e.g,
\begin{equation}
   L_{L2rec} = \| x - \tilde{x} \|_2,  \qquad
   L_{L2M} = \| x_m - \tilde{x}_m  \|_2,
\end{equation}
where $L_{L2rec}$ denotes the reconstruction loss at input-level and $L_{L2M}$ denotes the middle reconstruction loss. In our notation, $x$ represents the input (no subscripts) and $x_i$ (with subscripts) represent the feature map activations of the Convnet, respectively. Similarly, $\tilde{x}$ and $\tilde{x}_m$ are the input and activations of the Deconvnet, respectively.
The entire model architecture is shown in figure \ref{fig:model_pool}(b).
Notice in the following, we may use $L_{L2*}$ to represent the weighted sum of $L_{L2rec}$ and $L_{L2M}$.

\subsection{Soft version ``what'' and ``where''} 
\label{sub:soft}
Recently, \cite{goroshin2015learning} introduces a soft version of \emph{max} and \emph{argmax} operators within each pooling region:
\begin{align}
m_k   &= \sum\limits_{N_k} z(x,y) \frac{e^{\beta z(x,y)}}{\sum\limits_{N_k}  e^{\beta z(x,y)} } \approx \max\limits_{N_k} z(x,y) \\
\vec{p_k} &= \sum\limits_{N_k}
	\begin{bmatrix}
		x \\
		y
	\end{bmatrix}
	\frac{e^{\beta z(x,y)}}{\sum\limits_{N_k} e^{\beta z(x,y)} } \approx
	\argmax\limits_{N_k} z(x,y),
\end{align}
where $z(x,y)$ denotes activation on the feature maps and $x,y$ represent spatial location which take normalized values from -1 to 1. $N_k$ stands for the $k^{th}$ pooling region. Note that $\beta$ is a hyper-parameter that is always set to be non-negative. It parametrizes soft pooling in such a way that the larger the $\beta$, the closer the soft-pooling approaches max-pooling, while small $\beta$ approximates mean-pooling. We use interpolation in the unpooling stage to handle continuous value conveyed by ``where''.

The soft pooling and unpooling can be embedded seamlessly into the SWWAE model and it has the virtue such that it can backpropogate through $\vec{p}$, in the contrast to the hard max-pooling being not differentiable w.r.t the \emph{argmax} ``switch'' locations.
Furthermore, soft-pooling operators enable location information to be more accurately represented and thus enable the features to capture fine details about the input, as evidenced in our visualization experiments (see section \ref{sub:capsule}).


\subsection{Training with joint losses and regularization} 
\label{sub:training_with_joint_losses}
As we mentioned, the SWWAE provides a unified framework for learning with all three learning modalities, all within a single architecture and single learning algorithm, i.e. stochastic gradient descent and backprop. Switching between these modalities can be achieved as follows:
\begin{tightitemize}
\setlength\itemsep{1em}
\item for supervised learning, we can mask out the entire Deconvnet pathway by setting $\lambda_{L2*}$ to $0$ and the SWWAE falls back to vanilla Convnet. 
\item for unsupervised learning, we can nullify the fully-connected layers on top of Convnet together with softmax classifier by setting $\lambda_{NLL}=0$. In this setting, the SWWAE is equivalent to a deep convolutional auto-encoder.
\item for semi-supervised learning, all three terms of the loss are active. The gradient contributions from the Deconvnet can be interpreted as an information preserving regularizer.
\end{tightitemize}

The idea behind using reconstruction as a regularizer was studied previously in \cite{erhan2010does}, although it uses unsupervised pre-training as its setup.
In terms of this, SWWAE is connected to unsupervised pre-training in the sense that both paradigms attempt to provide better generalization by forcing the model to reconstruct. One argument of unsupervised learning acting as a regularizer is that supervised loss drives to model $P(Y \mid X)$, while unsupervised pre-training captures the input distribution of $P(X)$; and learning $P(X)$ is helpful to learning $P(Y \mid X)$ (\cite{erhan2010does}). However, we argue that applying this statement to unsupervised pre-training setup appears unconvincing. One can argue that using $P(X)$ merely to initialize the model for learning $P( Y \mid X)$ has a very weak effect; i.e. the gradients from learning $P( Y \mid X)$ completely overwrite the initial weights, thus eliminating any regularizing effect that may have been obtained from learning $P(X)$.
We argue that joint training is a more effective strategy, i.e. SWWAE; our approach tries to model $P(Y \mid X)$ together with $P(X)$ jointly during training. Comparisons between different regularizers are shown in appendix.

Moreover, training jointly with multiple losses helps avoid collapsing or learning trivial representation. For one thing, a common issue with auto-encoders is that they learn little more than the identity function; e.g, copying input to get perfect reconstruction. For another, sparse auto-encoders (\cite{makhzani2014winner}, \cite{makhzani2013k}) attain a well known trivial solutions: adding an $L_1$ penalty on the hidden layers is likely to scale down the encoder weights and scale up the decoders weight in order to reconstruct while achieving small activations. We argue that a direct way to avoid such trivial solutions is to include a supervised loss, which directly optimizes a non-trivial, useful, criterion that helps factorize the data into semantically relevant factors of variation.

\subsection{Intermediate L2 constraints}
\label{sub:l2m}
The reasons for adding intermediate L2 reconstruction terms are listed as follow. First, it prevents the feature planes from being shuffled so that the ``where'' map conveyed from encoder $i^{th}$ are guaranteed to match the ``what'' from decoder $i^{th}$. Otherwise, the unpooling may see ``what'' and ``where'' with shuffle orders, and hence cannot work properly. Second, in particular when training with classification loss, intermediate terms disallow the scenario that upper layers become idle while only lower layers are busy at reconstructing, in which case filters from those unemployed layers are not regularized. The related classification performance comparison about intermediate L2 terms is shown in appendix.
Third, As a correspondence to layer-wise auto-encoder training, each intermediate encoder/pool/unpool/decoder units in SWWAE, combined with intermediate L2 terms, can be seen as a single-layer convolutional auto-encoder (\cite{masci2011stacked}).


\section{Experiments}
\label{sec:exp}
We use the following notation to describe our architecture (assume square kernels) e.g. {\small \texttt{(16)5c-(32)3c-2p-10fc}}, in which `{\small \texttt{(16)5c}}' denotes convolution layer with 16 feature maps while kernel size being set to 5. {\small \texttt{2p}} denotes $2 \times 2$ pooling layer and {\small \texttt{10fc}} denotes fully-connection layer that connects to 10 hidden units. ReLU is omitted in the notation.

\subsection{Necessity of ``where''} 
\label{sub:necessity_of_where_}
\begin{figure}[h]
\centering
\minipage{0.4\textwidth}
\includegraphics[width=\linewidth]{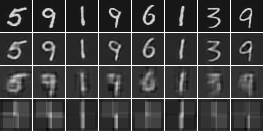}
\endminipage \hspace{10pt}
\minipage{0.4\textwidth}
\includegraphics[width=\linewidth]{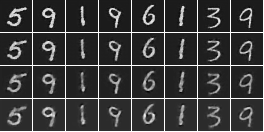}
\endminipage
\caption{Generation quality comparison between using upsampling (left) and unpooling (right). From top to bottom, the pooling sizes are respectively 2, 4, 8, 16.}
\label{fig:upsample}
\end{figure}

We address the necessity of ``where'' by showing the difference of reconstructions using ``where'' versus not using ``where''. Upsampling is an alternative way to do unpooling but without dreaming up ``where'', in the respect that ``what'' is agnostic about ``where'' and hence it gets copied on all the positions.
Figure \ref{fig:upsample} displays a group of reconstructed digits sampled from MNIST's testing set which are generated by a trained SWWAE using MNIST training set.
The architecture we use is: {\small \texttt{(16)5c-(32)3c-Xp}} and the pooling size being experimented varies from 2 to 16. Note we use hard max-pooling for this experiment and the architecture is trained in unsupervised mode.

On one hand, as the generations given by unpooling are obviously clearer and cleaner than the ones by upsampling, this experiment demonstrates that ``where'' is critical information demanded by reconstructing; one can barely obtain well reconstructed images without preserving ``where''.
On the other hand, this experiment can also be considered as an example using SWWAE for generative purpose.

\subsection{Invariance and Equivariance}
\label{sub:capsule}
\begin{figure}[h]
\centering
\minipage{0.25\textwidth}
\includegraphics[width=\linewidth]{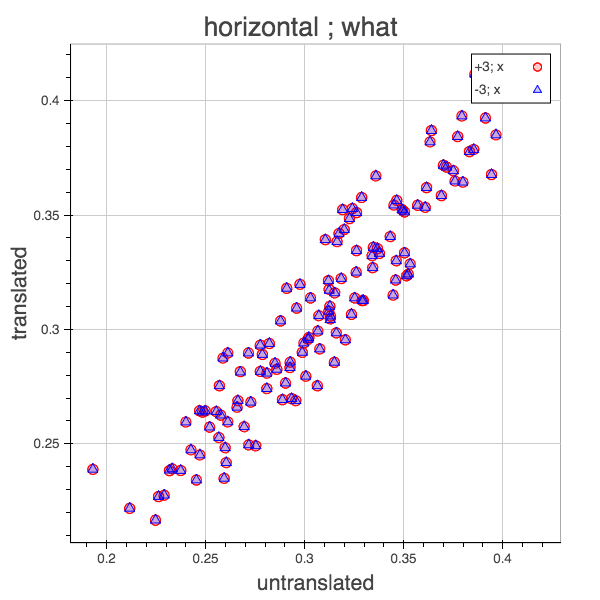}
\endminipage
\minipage{0.25\textwidth}
\includegraphics[width=\linewidth]{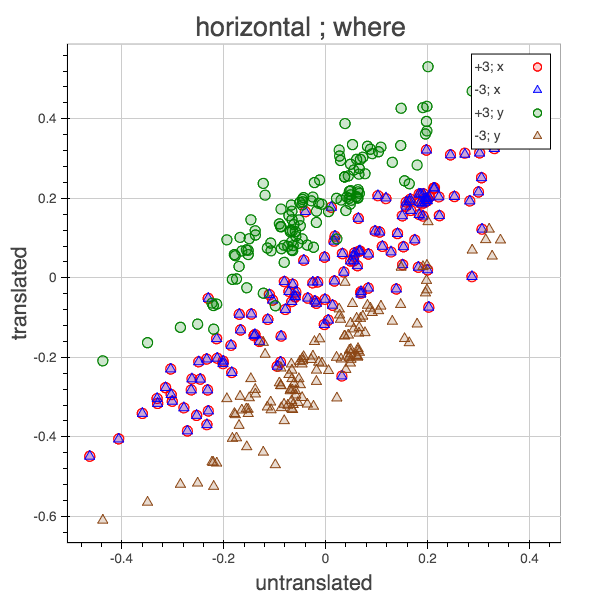}
\endminipage
\minipage{0.25\textwidth}
\includegraphics[width=\linewidth]{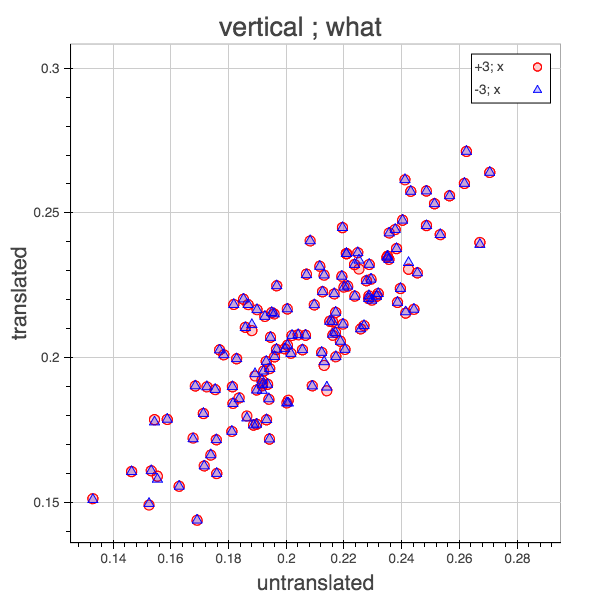}
\endminipage
\minipage{0.25\textwidth}
\includegraphics[width=\linewidth]{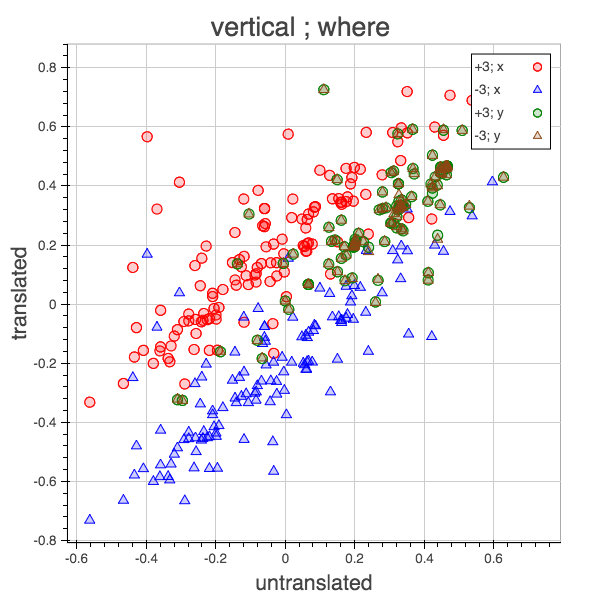}
\endminipage
\caption{Scatter plots depicting feature response produced by translating the input. The horizontal axis represents the ``what'' or ``where'' output from one feature plane for an untranslated digit image; vertical axis represents the ``what'' or ``where'' output from the same feature plane if that image is translated by +3 or -3 pixels in either horizontal or vertical direction.
From left to right, the figures are respectively: first (a): ``what'' of horizontally translated digits versus original digits;
second (b): ``where'' of horizontally translated digits versus original digits;
third  (c): ``what'' of vertically translated digits versus original digits;
fourth (d): ``where'' of vertically translated digits versus original digits.
Note that circles are used to feature +3 translation and triangles for -3.
In the ``where'' related plots, $x$ and $y$ denote two dimensions of ``where'' respectively.
}
\label{fig:capsule}
\end{figure}

\begin{figure}[h]
\centering
\includegraphics[width=0.8\linewidth]{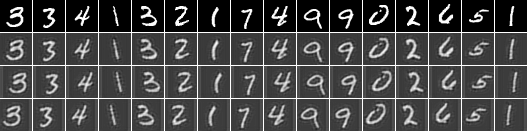}
\caption{Reconstructed MNIST digits in the capsule emulation experiments. The top row shows original input; second row shows the reconstruction of those original inputs; the bottom two rows display reconstruction of horizontally translated digits in positive and negative direction respectively.}
\label{fig:capsule_rec}
\end{figure}

In this section, we examine the relationship between ``what'' and ``where'' by using the visualization approach proposed with transforming auto-encoders (\cite{hinton2011transforming}) in which a number of ``capsules'' are trained to learn a representation consisting of equivariant and invariant components. Analogously, the ``what'' and ``where''  in our model's representation correspond to the invariant and equivariant components, respectively. The experiment recipe is stated as follow. (1) train a SWWAE using horizontally and vertically translated MNSIT digits from training set; (2) feed untranslated digits from testing set into SWWAE and obtain the ``what'' ($R$) and ``where'' ($R^2$); (3) horizontally or vertically translate same set of digits and feed it into SWWAE and cache ``what'' and ``where'' correspondingly; (4) plot the relationship between ``what'' and ``where'' obtained from translated digits versus untranslated ones, shown in figure \ref{fig:capsule}. The architecture we use is: {\small \texttt{(32)5c-(32)3c-2p-(32)3c-16p}} and we use soft pooling/unpooling with $\beta=100$. (5) since this experiment demands a large pooling size, we hence plot the generations in figure \ref{fig:capsule_rec} to make sure that SWWAE works appropriately under such large pooling settings.

%
%

We draw the conclusion from figure \ref{fig:capsule} that ``what'' and ``where'' behave much like the invariance and equivariance of capsules in \cite{hinton2011transforming}.
One one hand, ``where'' learns highly localized representation. Each element in the $R^2$ ``where'' has an approximately linear response to the pixel-level translation on either horizontal/vertical direction and learns to be invariant to another. On the other hand, ``what'' learns to be locally stable that exhibits strong invariance to the input-level translation.

\subsection{Classification performance}
\label{sec:class}
\begin{figure}[h]
\centering
\minipage{0.45\textwidth}
\includegraphics[width=\linewidth]{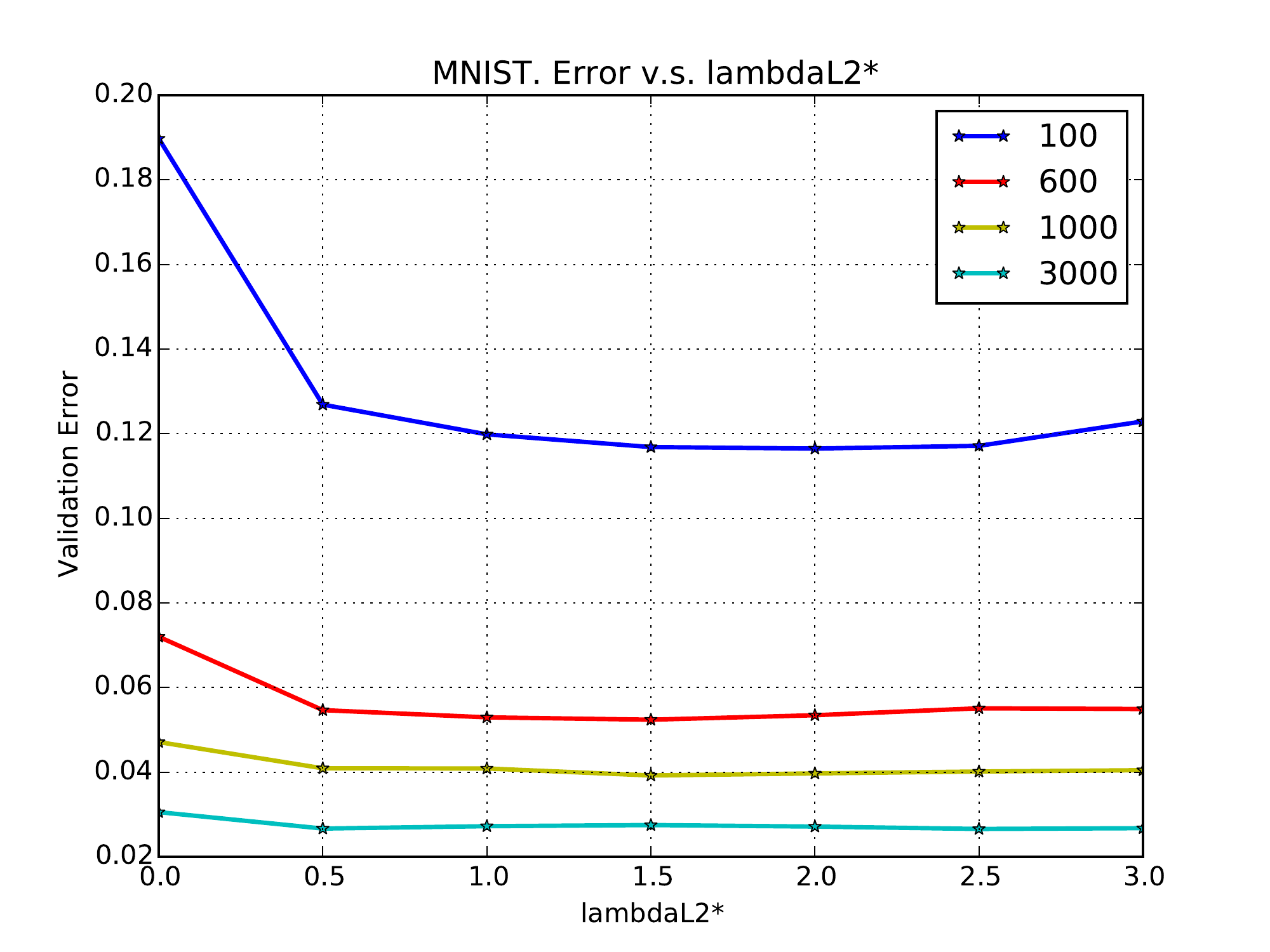}
\endminipage  \hspace{10pt}
\minipage{0.45\textwidth}
\includegraphics[width=\linewidth]{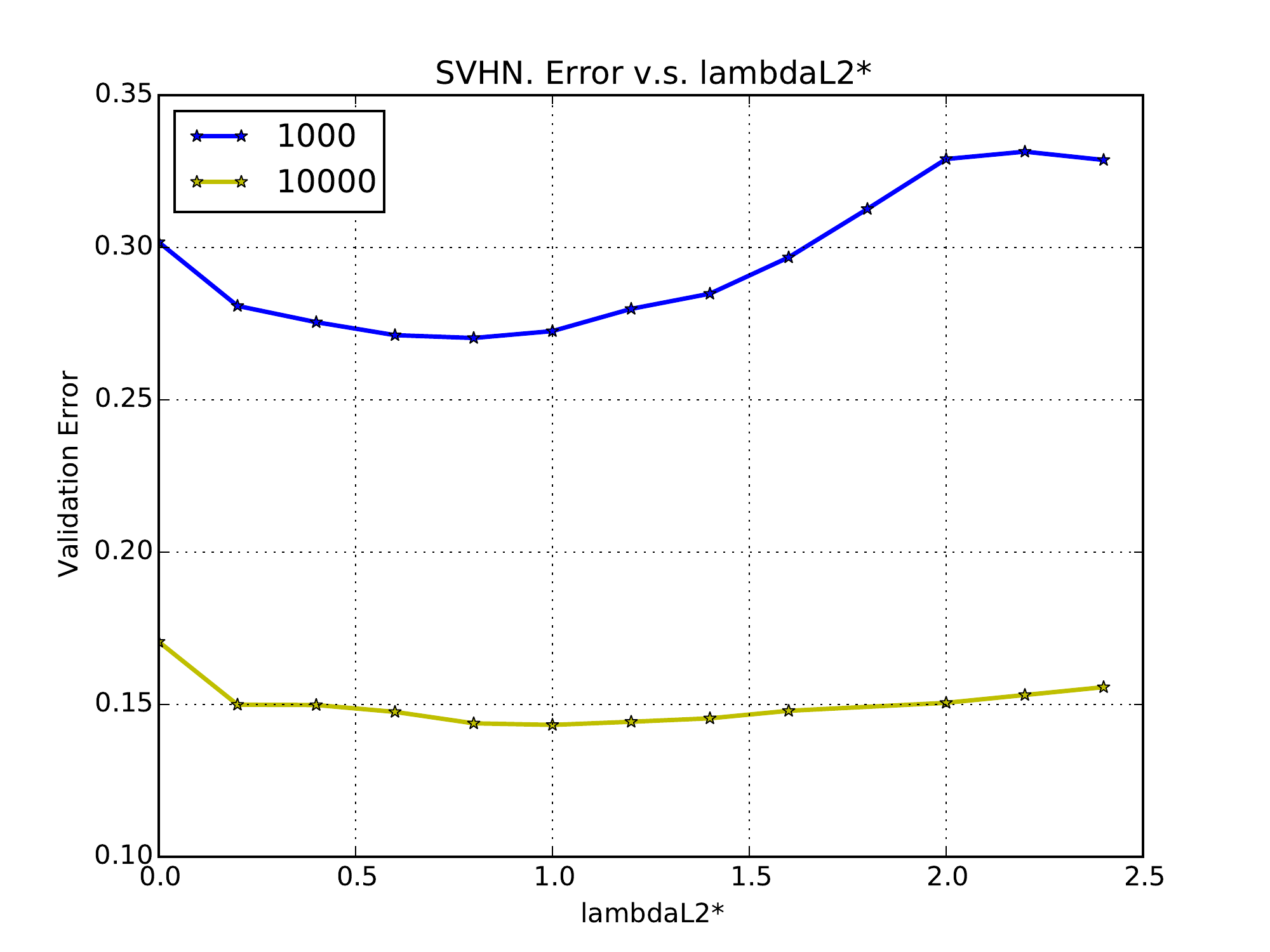}
\endminipage
\caption{Validation-error v.s. $\lambda_{L2*}$ on a range of datasets for SWWAE semi-supervised experiments. Left (a): MNIST. Right (b): SVHN. Different curves denote different number of labels being used.}
\label{fig:standalone}
\end{figure}

\subsubsection{MNIST \& SVHN}
\label{mnist_svhn}
As a start, we access the effect of SWWAE on classification by performing both semi-supervised and supervised experiments on MNIST and SVHN. We attempt to demonstrate that introducing a paired Deconvnet with a group of reconstruction losses can help generalization and provide an effective solution to make use of unlabeled data. Note in the classification experiments, we use the hard version pooling because it performs better than its soft counterparts in terms of classification.

We start by constructing semi-supervised datasets for both two datasets.
MNIST dataset consists of images of 10 different classes (0 to 9) of size 32x32 with 60,000 training samples and
10,000 test samples. We follow the previous work for data preparation: randomly select labeled samples from training set while the rest of the samples is used without labels The sizes of labeled subset are respectively 100, 600, 1000, 3000 and we ensure each class has same number of digits chosen in the labeled set.
SVHN dataset consists of 73,257 digits for training, 26,032 digits for testing and 53,1131 extra training samples that are less difficult.
Likewise, we construct labeled dataset for SVHN that contains 1000 samples uniformly distributed in 10 classes, chosen randomly from the non-extra training set.
In order to attain reliable results, we run each experiment several rounds whereby datasets are refreshed before each round and we average the performances of all rounds as the final evaluation.

We approach the ``standalone'' regularization effect of SWWAE on both datasets, by plotting the validation error v.s. $\lambda_{L2*}$ ($\lambda_{L2M}$ and $\lambda_{L2rec}$ are combined to be equal for this experiment) in figure \ref{fig:standalone}. By ``standalone'', we mean that no other well-known regularizer is applied.

We further evaluate SWWAE on the testing set of SVHN with the chosen hyper-parameters indicated by validation error. Table \ref{tab:svhn} shows the results.
We additionally evaluate SWWAE on SVHN under pure supervised manner (with all the available labels) that we find that the testing error decreases from \textbf{5.89\%} to \textbf{4.94\%} yielded by SWWAE versus a vanilla Convnet under same configuration.
The architecture we use for MNIST and SVHN are respectively {\small \texttt{(64)5c-2p-(64)3c-2p-(64)3c-2p-10fc}} and {\small \texttt{(128)5c-2p-(128)3c-(256)3c-2p-(256)3c-2p-10fc}}.
More exploration on MNIST is shown in appendix.

\begin{table}[t]
  \caption{Comparison between SWWAE and other best published results on SVHN with 1000 labels.}
\label{tab:svhn}
    \centering
    \small
\begin{tabular}{ll}
\multicolumn{1}{c}{\bf model / N} &\multicolumn{1}{c}{\bf error rate (in \%)}
\\ \hline \noalign{\vskip 1mm}
KNN & $77.93$ \\
TSVM (\cite{vapnik1998statistical}) & $66.55$  \\
M1+KNN (\cite{kingma2014semi}) & $65.63$ \\
M1+TSVM (\cite{kingma2014semi})& $54.33$ \\
M1+M2 (\cite{kingma2014semi}) & $36.02$
\\ \hline \hline
SWWAE without dropout ($\lambda_{L2*}=0.8$)& $27.83$   \\
SWWAE with dropout ($\lambda_{L2*}=0.4$) & $\mathbf{23.56}$   \\
\end{tabular}
\end{table}

\subsubsection{STL-10}
STL-10 contains larger 96x96 pixel images and relatively less labeled data (5000 training samples, 100,000 unlabeled samples and 8,000 test samples). The training set is mapped to 10 predefined folds with 1,000 images each. Therefore, STL-10 has a 100:1 ratio of the amount of unlabeled samples to the labeled ones in each fold.
We follow the testing protocol of STL-10 that we first tune the hyper-parameters for each fold by validation error and let the best performed model predict the testing set. The final score is reported by averaging the testing score of 10 folds.
For STL-10, we access the possibility to combine batch normalization (\cite{ioffe2015batch}) and SWWAE. Furthermore, we carry out spatial batch normalization which preserves the mean and standard deviation from each feature map while they get normalized independently based on their own statistics.
We devise a VGG-style (\cite{simonyan2014very}) deep net, {\small \texttt{(64)3c-4p-(64)3c-3p-(128)3c-(128)3c-2p-(256)3c-(256)3c-(256)3c-(512)3c-\\(512)3c-(512)3c-2p-10fc}} and each convolution layer is followed by a spatial batch normalization layer, which is applied in both Convnet and Deconvnet pathways. Results are shown in table \ref{tab:stl-10}.

\begin{table}[t]
\caption{Comparison between SWWAE and other best published results on STL-10.}
\label{tab:stl-10}
\centering
\small
\begin{tabular}{ll}
\multicolumn{1}{c}{\bf model} &\multicolumn{1}{c}{\bf accuracy}
\\ \hline \noalign{\vskip 1mm}  
Multi-task Bayesian Optimization (\cite{swersky2013multi}) & $70.1\%$ \\
Zero-bias Convnets + ADCU (\cite{paine2014analysis}) & $70.2\%$ \\
Exemplar Convnets (\cite{dosovitskiy2014discriminative}) & $\mathbf{75.4\%}$
\\ \hline \hline
SWWAE & $74.33\%$  \\
Convnet of same configuration & $57.45\%$
\end{tabular}
\end{table}

\subsection{Large scale experiments} 
\label{sub:large_scale_experiments}

\subsubsection{CIFAR with 80 million tiny images}
\label{cifar}

The dataset CIFAR-10 and CIFAR-100 are sampled and labeled from the 80 million tiny images dataset (\cite{torralba200880}). Both datasets contain 60,000 32x32 images which are small portions of the set of 80 million images.
In contrast to the former classification experiments, this experiment involves substantially more abundant unlabeled data in relation to the amount of labeled data. We carry out the SWWAE with a VGG-style network (\cite{simonyan2014very}): {\small \texttt{(128)3c-(256)3c-2p-(256)3c-(512)3c-2p-(512)3c-(512)3c-2p-(512)3c-(512)3c-\\2p-128fc-10fc}} in which each convolution is bundled and followed by spatial batch normalization (\cite{ioffe2015batch}) in both Convnet and Deconvnet. To compare with results from other approaches, we perform the experiments in the common experimental setting that only adopts contrast normalization, small translation and horizontal mirroring for data preprocessing. The results are shown in table \ref{tab:cifar}.

\begin{table}[t]
\centering
\small
\caption{Accuracy of SWWAE on CIFAR-10 and CIFAR-100 in comparison with best published single-model results. Our results are obtained with the common experimental setting that we only adopt contrast normalization, small translation and horizontal mirroring for data preprocessing.}
\label{tab:cifar}
\begin{tabular}{lll}
\multicolumn{1}{c}{\bf model} & \multicolumn{1}{c}{\bf CIFAR-10} & \multicolumn{1}{c}{\bf CIFAR-100}
\\ \hline \noalign{\vskip 1mm} 
All-Convnet (\cite{springenberg2014striving}) & $\mathbf{92.75\%}$ & $66.29\%$ \\
Highway Network (\cite{srivastava2015training}) & $92.40\%$ & $67.76\%$ \\
Deeply-supervised nets (\cite{lee2014deeply}) & $92.03\%$ & $65.43\%$ \\
Fractional Max-pooling with large augmentation (\cite{graham2014fractional}) & $95.50\%$ & $68.55\%$
\\ \hline \hline
SWWAE ($\lambda_{L2rec}=1, \lambda_{L2M}=0.2$) & $92.23\%$ & $\mathbf{69.12\%}$  \\
Convnet of same configuration & $91.33\%$ & $67.50\%$ 
\end{tabular}
\end{table}


\section{Conclusion and outlook}
\label{sec:concl}
The overall system, which can be seen as pairing a Convnet with a Deconvnet, yields good accuracy on a variety of semi-supervised and supervised tasks.
We envision that such architecture may also be useful in video related tasks where unlabeled samples abound.

\section*{Acknowledgments}
We thank Xiang Zhang and Aditya Ramesh for many useful discussions.

{\small
\bibliography{iclr2016_workshop}
\bibliographystyle{iclr2016_workshop}
}

\newpage
\section*{Appendix: more MNIST} 
\label{app_mnist}

We tend to exhibit more experimental results on MNIST in two respects.
First, on the validation set, we compare the performance of SWWAE against other regularization methods, shown in table \ref{tab:mnist_val}. Note in order to make the comparison more realistic and closer to practical uses, we add dropout (\cite{hinton2012improving}) at fully-connected layers as the default for this set of comparisons. Regularizers under comparison include dropout on the convolution layers and L1 sparsity penalty on hidden layers.
Besides, we also train SWWAE unsupervisedly and separately train a softmax classifier afterwards using labeled samples; this disjointly trained architecture is denoted by ``unsup-sfx''. We similarly try using SWWAE as an unsupervised pre-training approach, followed by fine-tuning the entire Convnet part driven by labeled data, which is denoted by ``unsup-pretr''. Note the difference between ``unsup-pretr'' and ``unsup-sfx'' lies in if the Convnet part is frozen when training the softmax classifier on top.
In addition, ``noL2M'' is written for experiments that SWWAE is trained with only reconstruction loss at the input level, i.e. $\lambda_{L2M}=0$ and $\lambda_{L2rec}$ is chosen by validation error.
Second, we report the testing set error rate obtained by SWWAE with chosen hyper-parameter of SWWAE and compare it with best published results in table \ref{tab:mnist_test}.
Note that for the experiments on MNIST testing set, the labeled set is generated by sampling from the entire MNIST training set; the experiments on validation set, instead, sample the labeled data only from a subset of the MNIST training set because the rest of which is deemed as validation set.
The SWWAE configuration is {\small \texttt{(64)5c-2p-(64)3c-2p-(64)3c-2p-10fc}}.

\begin{table}[h]
\caption{Comparison against other regularization approaches and disjoint training approaches on MNIST dataset. The scores are validation error rate (in \%). Dropout is added at the fully-connected layers as default.}
  \centering
  \small
\begin{tabular}{lllll}
\multicolumn{1}{c}{\bf model / N} &\multicolumn{1}{c}{\bf 100} &\multicolumn{1}{c}{\bf 600} &\multicolumn{1}{c}{\bf 1000} &\multicolumn{1}{c}{\bf 3000}
\\ \hline \noalign{\vskip 1mm}
SWWAE & $\mathbf{10.66\pm0.55}$ & $\mathbf{4.35\pm0.30}$ & $3.17\pm0.17$ & $2.13\pm0.10$ \\
dropout on convolution  & $14.23\pm0.94$ & $4.70\pm0.38$ & $3.37\pm0.11$ & $2.08\pm0.10$ \\
L1 & $10.91\pm0.29$ & $4.61\pm0.28$ & $3.55\pm0.31$ & $2.67\pm0.25$ \\
unsup-sfx  & $17.81\pm0.06$ & $8.41\pm0.08$ & $6.40\pm0.06$ & $4.76\pm0.03$ \\
unsup-pretr  & - & $9.80\pm0.06$ & $6.135\pm0.03$ & $4.41\pm3.11$ \\
noL2M  & $12.41\pm1.95$ & $4.63\pm0.24$ & $\mathbf{3.15\pm0.22}$ & $\mathbf{2.08\pm0.18}$
\end{tabular}
\label{tab:mnist_val}
\end{table}

\begin{table}[h]
\caption{Comparison of testing error rate (in \%) between SWWAE and other best published results on MNIST dataset within semi-supervised setting.}
\label{tab:mnist_test}
 \centering
 \small
\begin{tabular}{lllll}
\multicolumn{1}{c}{\bf model / N} &\multicolumn{1}{c}{\bf 100} &\multicolumn{1}{c}{\bf 600} &\multicolumn{1}{c}{\bf 1000} &\multicolumn{1}{c}{\bf 3000}
\\ \hline \noalign{\vskip 1mm}
Convnet (\cite{lecun1998gradient}) & $22.98$ & $7.86$ & $6.45$ & $3.35$ \\
TSVM (\cite{vapnik1998statistical}) & $16.81$ & $6.16$ & $5.38$ & $3.45$ \\
CAE (\cite{rifai2011contractive}) & $13.47$ & $6.3$ & $4.77$ & $3.22$ \\
MTC (\cite{rifai2011manifold}) & $12.03$ & $5.13$ & $3.64$ & $2.57$ \\
PL-DAE (\cite{lee2013pseudo}) & $10.49$ & $5.03$ & $3.46$ & $2.69$ \\
WTA-AE (\cite{makhzani2014winner}) & - & $\mathbf{2.37}$ & $1.92$ & - \\
M1+M2 (\cite{kingma2014semi}) & $3.33\pm0.14$ & $2.59\pm0.05$ & $2.40\pm0.02$ & $2.18\pm0.04$ \\
LadderNetwork (\cite{rasmus2015semi}) & $\mathbf{1.06\pm0.37}$ & - & $\mathbf{0.84\pm0.08}$ & - \\
\hline \hline
SWWAE without dropout & $9.17\pm0.11$ & $4.16\pm0.11$ & $3.39\pm0.01$ & $2.50\pm0.01$ \\
SWWAE with dropout & $8.71\pm0.34$ & $3.31\pm0.40$ & $2.83\pm0.10$ & $\mathbf{2.10\pm0.22}$
\end{tabular}
\end{table}

Aside from semi-supervised setting, we also explore SWWAE training on full labeled training dataset in which we find that SWWAE achieves a better testing error rate \textbf{0.71\%} versus \textbf{0.76\%} obtained by Convnet under same configuration.

We reason that SWWAE not working so well as Ladder networks \cite{rasmus2015semi} is due to the fact that reconstructing MNIST digits is overly easy for SWWAE. 
Assume we have an one layer SWWAE with one pooling and unpooling layer implemented into two pathway respectively.
Since MNIST is a roughly binary dataset (0/1) and thus within unpooling stage, decoding doesn't necessarily demand the information from ``what'' for reconstruction; i.e., it could get perfect reconstruction by pinning $1$ on the positions indicated by ``where''. Therefore, we believe that reconstructing MNIST dataset renders insufficient regularization on the encoding pathway. However, this phenomenon won't happen on other natural image datasets, such as CIFAR or STL-10 where we show good results by SWWAE.


\end{document}